\documentclass[10pt,twocolumn,letterpaper]{article}

\usepackage{times}
\usepackage{graphicx}
\usepackage{amsmath}
\usepackage{amssymb}
\usepackage{dsfont}
\usepackage{multirow}
\usepackage{authblk}

\usepackage[pagebackref=true,breaklinks=true,letterpaper=true,colorlinks,bookmarks=false]{hyperref}

\begin{document}

\title{Contour Loss: Boundary-Aware Learning for Salient Object Segmentation}

\author{Zixuan Chen}
\author{Huajun Zhou}
\author{Xiaohua Xie\thanks{Corresponding author}}
\author{Jianhuang Lai}
\affil[]{Sun Yat-sen University}
\affil[]{xiexiaoh6@mail.sysu.edu.cn}

\renewcommand\Authfont{\large}
\renewcommand\Affilfont{\small}

\date{}
\maketitle

\begin{abstract}

We present a learning model that makes full use of boundary information for salient object segmentation. Specifically, we come up with a novel loss function, i.e., Contour Loss, which leverages object contours to guide models to perceive salient object boundaries. Such a boundary-aware network can learn boundary-wise distinctions between salient objects and background, hence effectively facilitating the saliency detection. Yet the Contour Loss emphasizes on the local saliency. We further propose the hierarchical global attention module (HGAM), which forces the model hierarchically attend to global contexts, thus captures the global visual saliency. Comprehensive experiments on six benchmark datasets show that our method achieves superior performance over state-of-the-art ones. Moreover, our model has a real-time speed of 26 fps on a TITAN X GPU.

\end{abstract}

\section{Introduction}
Salient object segmentation, which aims to extract the most conspicuous object regions in visual range, has become an attractive computer vision research topic over the decades.
A vast family of saliency algorithms have been proposed to tackle the saliency segmentation problem, which distinguishes whether a pixel pertains to a noticeable object or inconsequential background.
Due to a mixture of both object and background, pixels closed to the boundary between object and background are error-prone.

\begin{figure}[!t]
\centering
\begin{minipage}{0.115 \textwidth}
\includegraphics[width=0.81in,height=0.65in]{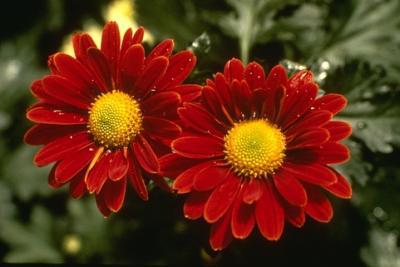}
\end{minipage}
\begin{minipage}{0.115 \textwidth}
\includegraphics[width=0.81in,height=0.65in]{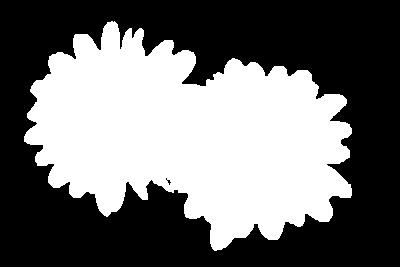}
\end{minipage}
\begin{minipage}{0.115 \textwidth}
\includegraphics[width=0.81in,height=0.65in]{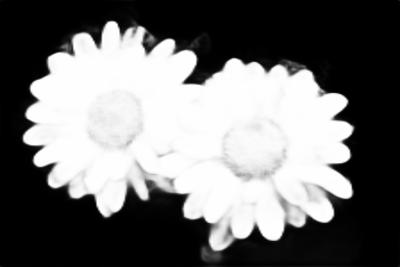}
\end{minipage}
\begin{minipage}{0.115 \textwidth}
\includegraphics[width=0.81in,height=0.65in]{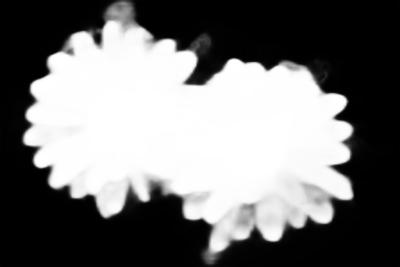}
\end{minipage}

\begin{minipage}{0.115 \textwidth}
\includegraphics[width=0.81in,height=0.65in]{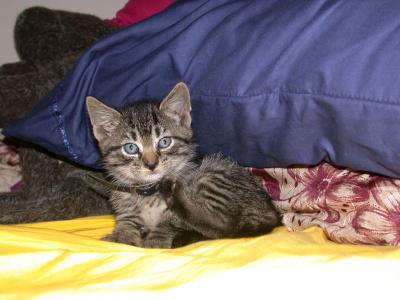}
\end{minipage}
\begin{minipage}{0.115 \textwidth}
\includegraphics[width=0.81in,height=0.65in]{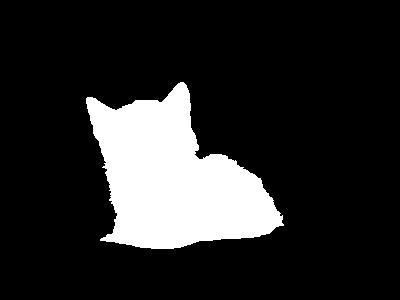}
\end{minipage}
\begin{minipage}{0.115 \textwidth}
\includegraphics[width=0.81in,height=0.65in]{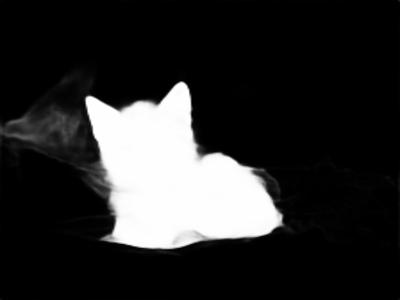}
\end{minipage}
\begin{minipage}{0.115 \textwidth}
\includegraphics[width=0.81in,height=0.65in]{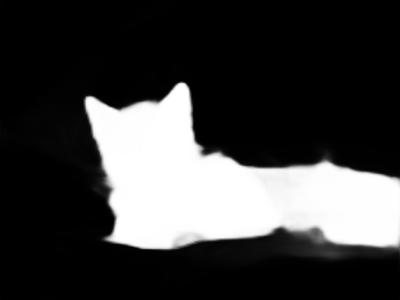}
\end{minipage}

\begin{minipage}{0.115 \textwidth}
\includegraphics[width=0.81in,height=0.65in]{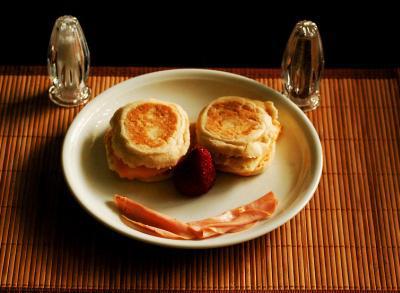}
\centering (a)
\end{minipage}
\begin{minipage}{0.115 \textwidth}
\includegraphics[width=0.81in,height=0.65in]{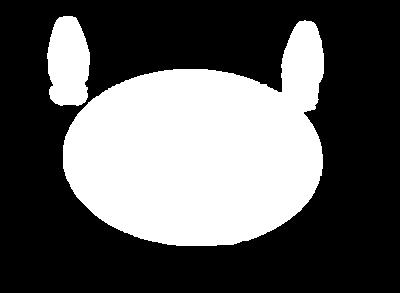}
\centering (b)
\end{minipage}
\begin{minipage}{0.115 \textwidth}
\includegraphics[width=0.81in,height=0.65in]{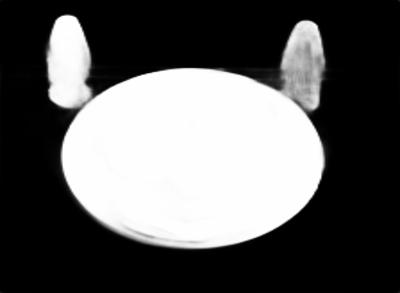}
\centering (c)
\end{minipage}
\begin{minipage}{0.115 \textwidth}
\includegraphics[width=0.81in,height=0.65in]{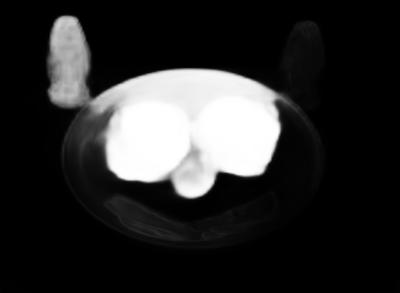}
\centering (d)
\end{minipage}
\vspace{2pt}

\caption{Visual examples of the proposed method and PCA \cite{PiCANet}. 
From left to right: (a) original image, (b) ground truth, (c) ours, (d) PCA \cite{PiCANet}
}
\label{fig:examples}
\end{figure}

Early methods \cite{1998model,CMM,DUT-OMRON} determine the saliency by utilizing hand-crafted appearance features.
These methods often focus on the low-level visual features and are struggling to achieve satisfactory results in the image with a complex scene. 
Compared with the methods based on hand-crafted features and prior knowledge, the fully convolution network (FCN) based frameworks \cite{FCN,U-Net,FPN} made remarkable progress by exploiting high-level semantic information.
To learn the binarized saliency segmentation, these networks usually adopt cross entropy loss as objective function.
However, cross entropy loss only considers sample distribution while neglecting the appearance cues of target objects, such as boundaries and inner textures.
Especially, the boundary is regarded very important for saliency segmentation.

To leverage the boundary information, multi-task architecture \cite{NLDF,CKT} was designed to aggregate the features acquired from boundary and saliency labels.
As the supervised information is distinct between boundary and saliency branches, simply aggregating these features may lead to incompatible interference, thus their models are hard to converge.
Since aforementioned saliency frameworks cannot well exploit boundary information to determine the contour pixels, these methods may obtain a sub-optimal result with vague boundaries.

Because the final decision of salient object regions relies on the spatial contexts, exploiting contextual information in models can lessen the misdirections of insignificant background.
Recently, attention mechanism was exploited to obtain attended features for capturing global contexts by \cite{PiCANet,BRN,PAGRN}.
However, as these attended features are generated by softmax in global forms, they only emphasize several significant pixels and abnegate other information in images.
Therefore, for the high-resolution image, it is not a good choice to capture the global contexts by softmax-wise attention modules, which easily leads to overfitting in training.

To address aforementioned issues of boundary-aware learning and attention mechanism for saliency segmentation, we propose a novel segmentation loss and an effective global attention module, i.e., Contour Loss and hierarchical global attention module (HGAM).
The aim of Contour Loss is to guide the network to perceive the object boundaries to learn the boundary-wise distinctions between salient objects and background.
Motivated by the focal loss \cite{focalloss}, we apply spatial weight maps in cross entropy loss, which assigns a relatively high value to emphasize the pixels near object borders in training.
As a result, the trained model is sensitive to the boundary-wise distinctions in images.
Since the Contour Loss focuses on local boundaries, HGAM is proposed to hierarchically attend to global contextual information for alleviating background distractions.
Different from the abovementioned attention modules which work with softmax, HGAM is based on global contrast thus can capture global contexts in all resolutions.
Our baseline model is based on FPN \cite{FPN} architecture with VGG-16 \cite{vgg} backbone, which is refined by employing residual blocks instead of simple convolution layers in decoder module.
With the help of the abovementioned techniques, our network yields state-of-the-art performance on six benchmarks.

Followings are the summary of our main contributions:

\vspace{0.1cm}
\textbf{1)}
We propose Contour Loss to guide networks to perceive salient object boundaries. 
Consequently, boundary-aware features can be obtained to facilitate final predictions on object boundaries.

\vspace{0.1cm}
\textbf{2)}
We present the hierarchical global attention module (HGAM) to attend to global contexts for reducing background distractions. 

\vspace{0.1cm}
\textbf{3)}
We construct a network based on FPN architecture and incorporate those proposed methods for joint training.

\vspace{0.1cm}
\textbf{4)}
Comprehensive experimental results and extensive in-depth analysis can explain the outperformance of proposed methods.
In addition, our model is very fast which has speed of 26 fps on an NVIDIA TITAN X GPU.

\section{Related Works}

\subsection{Salient Object Segmentation}
In recent years, various frameworks, including conventional methods and fully convolution network (FCN) based models, have been presented to address the problems of salient object segmentation.
We briefly review these two categories of methods in the followings.

\subsubsection{Conventional methods}
Conventional saliency detection methods utilize prior knowledge, as well as hand-crafted appearance features to capture salient regions.
Considering obvious distinctions between salient regions and background in pictures, local contrast is used to determine the pixel is conspicuous or not by \cite{LC}. 
Inspired by the effectiveness of local contrast, Cheng et al. \cite{CMM} proposed to capture the salient regions by global contrast.
To exploit different appearance cues for refining the saliency quality, multi-level segmentation model is designed by \cite{ECSSD,DRFI} to hierarchically aggregate these cues. 
As conventional methods only leverage low-level visual features, the lack of semantic information can lead to failures in complex situations.

\subsubsection{FCN-based method}
With the development of deep learning, remarkable progress has been made by FCN-based models.
Different from conventional methods, high-level semantic features can be exploited by FCNs to achieve better results.

\vspace{-0.4cm}
\paragraph{Typical FCNs.}
Long et al. \cite{FCN} first build a FCN for addressing semantic segmentation problems.
Ronneberger et al. \cite{U-Net} propose the U-Net architecture, which consists of a contracting path, a symmetric expanding path and lateral connections to integrate features with the same resolution.
To exploit the potential of deep feature pyramids, Lin et al. \cite{FPN} present the FPN architecture, which based on U-Net and employs hierarchical predictions.
These architectures are popularly followed by later related works.
\vspace{-0.4cm}
\paragraph{Recurrent structures.}
Inspired by RNNs, some recurrent structures have been proposed to tackle saliency segmentation problems.
Kuen et al. \cite{RACDNN} first design a recurrent network with convolution and deconvolution layers to enhance saliency maps from coarse to fine.
Liu and Han \cite{DHS} present an U-Net based architecture, which refines saliency maps by recursively integrating hierarchical predictions.
Wang et al. \cite{RFCN} utilize saliency results as feedback signals to improve saliency performance.
In \cite{Amulet}, Zhang et al. propose a complex recurrent structure to recursively extract and aggregate features.
Although these advanced recurrent structures can better leverage the potential of hierarchical features, the lack of heuristic knowledge limits their capability.

\vspace{-0.4cm}

\paragraph{Attention networks.}
The aim of attention mechanism is to adaptively select significant features, in other words, alleviating background distractions.
Since both attention and saliency have similar contextual meanings in pictures, recently many researchers adopt attention mechanism for saliency detection.
To alleviate background distractions, Wang et al. \cite{BRN} obtain attention maps from encoded features to attend to the global contexts, while Zhang et al. utilize both spatial and channel-wise attentions in \cite{PAGRN}.
Because softmax only emphasizes several pixels in image, softmax-wise attention modules are hard to capture global contexts in high-resolution.
To tackle this problem, Liu et al. \cite{PiCANet} propose global and local attention modules to capture global contexts and local contexts in low-resolution and high-resolution respectively, and Chen et al. \cite{RA} employ hierarchical predictions as attention maps, which can attend to global contexts in all resolutions.
As not all the features in background regions are helpless for saliency determination especially in deep layers, the predicted maps which are trained to close to annotation masks may lose some crucial information.
In contrast	to the aforementioned attention modules, the proposed HGAM can not only capture global contexts in all resolutions but also considers some crucial information from background regions.

\begin{figure*}[!t]
\centering
\includegraphics[width=6.5in,height=3in]{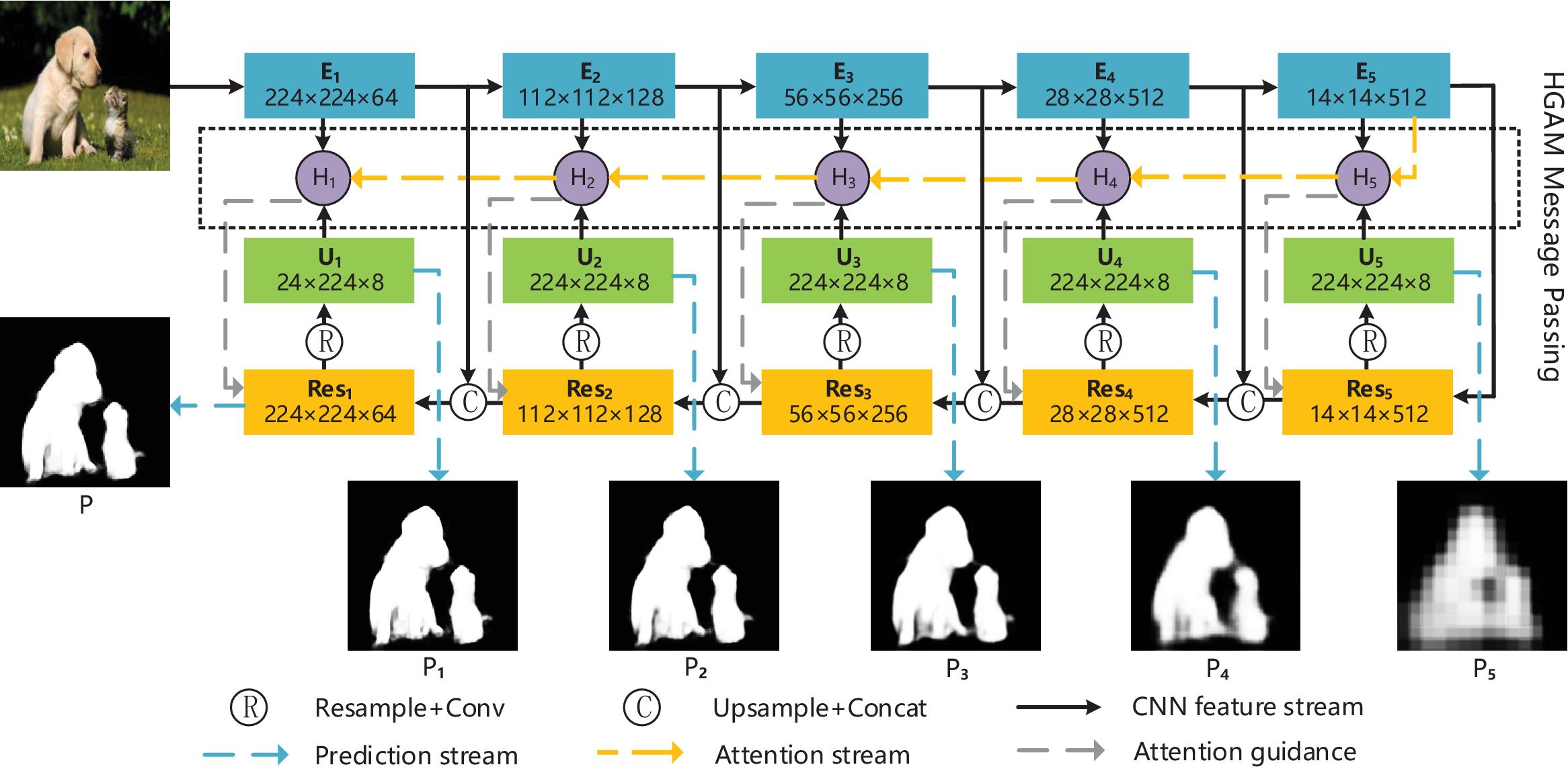}
\caption{Overall architecture of the proposed network with VGG-16 \cite{vgg} backbone. 
$E_{i}$ represents the feature of $i$th level in backbone.
$Res_{i}$ indicates the residual feature generated by the $i$th residual block.
$U_{i}$ denotes the $i$th resampled feature with $224\times224$ resolution, and $P_{i}$ is generated by $U_{i}$.
$H_{i}$ is the $i$th HGAM, which receives $E_{i}$, $U_{i}$ and previous HGAM message to guide $Res_{i}$.
$P$ denotes the final saliency output generated by guided $Res_{1}$.
}
\label{fig:baseline}
\end{figure*}

\subsection{Boundary-aware Learning}
One of the major challenges in saliency segmentation is to determine the conspicuous object boundaries.
Some researchers have pay attention to this point.

Since superpixel methods like SLIC \cite{SLIC} can obtain the regions by aggregating adjacent pixels with similar attributes, they are usually adopted to refine saliency results.
Yang et al. \cite{DUT-OMRON} propose a background-prior method, which utilize superpixel methods to obtain regions and detect salient regions by ranking the similarity of foreground or background units.
To revise the vague boundaries, \cite{DLS,DS,HKU-IS} employ superpixel algorithms to generate the object contours by these over-segmented regions.
Because superpixel relies on the distinction of pixel integration, it cannot well segment the pixels from low contrast regions.
Besides, these superpixel-based methods often have a huge computational cost.

Insteads of using superpixel methods for boundary determination, recent researches prefer to straightly leverage contour information in an entire framework.
As a conventional method, \cite{DSCD} build a two-stream framework for the mixture of texture and contour.
Luo et al. \cite{NLDF} and Xin et al. \cite{CKT} present a multi-task network architecture based on U-Net, which predicts both saliency and contour maps of the corresponding salient objects.
Due to the great distinctions between the saliency and boundary maps, it leads to inconsistent interference by simply aggregating these features.
Therefore, these models which are difficult to converge may generate sub-optimal results.

Different from abovementioned boundary-aware methods, the proposed Contour Loss can help the model to perceive the object boundaries by focusing on the boundary pixels, which is more robust and easier to be convergence.

\section{Proposed Methods}
Our proposed method mainly integrates a basic network with a Contour Loss and a hierarchical global attention module (HGAM), which aim at acquiring boundary-aware features and hierarchically integrating global contexts in all resolutions to enhance saliency results.
We describe our methods and baseline network in the following subsections.
The overall network structure is shown in Fig \ref{fig:baseline}.

\subsection{Baseline Network}
The FPN \cite{FPN} based baseline model without HGAMs and $P$ is shown in Fig \ref{fig:baseline}.
The network mainly consists of two categories of modules: encoder module and decoder module.

For encoder module, we adopt the VGG-16 \cite{vgg} backbone which is pretrained on ImageNet \cite{imagenet} for image classification. 
As the resolution of input image $I$ is $224\times224$, to adapt the saliency segmentation task, we utilize the backbone to extract feature maps at 5 levels, which can be represented as encoded features $F^{E}=\{E_{i},1\le i\le 5 \}$ with the resolution $w_{i}\times h_{i} = \frac{224}{2^{i-1}}\times \frac{224}{2^{i-1}}$.
Since $F^{E}$ are extracted at multi-levels, they contain both low-level visual cues and high-level semantic information from different resolutions.
To integrate these multi-level information, we transfer $F^{E}$ to decoder module.

Because residual block is better than pure convolution layer in aggregating the multi-scale features, our decoder module is constructed by 5 residual blocks corresponding to $F^{E}$.
After the decoder module has received $F^{E}$, it generates the residual features $F^{R} = \{Res_{i},1\le i\le 5 \}$ and each $Res_{i}$ can be formulated as:
\begin{equation}
Res_{i}=
\begin{cases}
\delta(\{Res_{i+1}\}^{up\times2}\oplus E_{i}; \theta^{R}_{i}) & 1\le i<5 \\
\delta(E_{i}; \theta^{R}_{i}) & i=5
\end{cases}	
\end{equation}
where $\delta(\star;\theta)$ stands for the convolution together with ReLU layers with parameters $\theta=\{W, b\}$.
$\oplus$ and $\{\star\}^{up\times2}$ represent the channel-wise concatenation and the upsample operation by a factor 2 respectively.
To achieve the hierarchical predictions like FPN, we resample $F^{R}$ to $224\times224$ resolution for obtaining the upsampled features $F^{U}=\{U_{i},1\le i\le 5 \}$, then utilize these feature maps to generate the hierarchical predictions $F^{P}=\{P_{i},1\le i\le 5 \}$.
$U_{i}$ and $P_{i}$ can be formulated as:
\begin{equation}
\begin{split}
U_{i} &= \delta(\{Res_{i}\}^{up\cdot224}, \theta^{U}_{i}) \\
P_{i} &= \eta(U_{i}, \theta^{P}_{i})
\end{split}
\end{equation}
where $\{\star\}^{up\cdot224}$ denotes upsampling features to $224\times224$ resolution and $\eta(\star;\theta)$ stands for the convolution together with the Sigmoid layers with parameters $\theta=\{W, b\}$.

As the $P_{5}$, ..., $P_{1}$ are based on $F^{R}$ from low to high resolutions, these prediction maps can receive various supervised information from coarse to fine.
To better leverage these various feedbacks from loss for updating parameters, in the training phase, the loss is calculated by the weighted sum of $F^{P}$ like \cite{DSS}, it can be formulated as:
\begin{equation}
Loss(F^{P};Y)=\sum_{i=1}^{5}{W_{i}^{L}\cdot loss(P_{i};Y)}
\label{eq:loss}
\end{equation}
where $Y$ is the annotation mask, the $loss(\star;Y)$ and $Loss(\star;Y)$ represent the cross entropy loss and its weighted combination respectively.
$W_{i}^{L}$ is the hyperparameter of corresponding prediction $P_{i}$.
In testing, we adopt the $P_{1}$ as saliency result.

\subsection{Contour Loss}
Salient object segmentation aims at capturing the most conspicuous objects in input images.
Suppose images only contain two parts: the background and salient objects.
For most pixels, they locate at the inside of the objects or background, which indicate that they are far from the object borders.
Intuitively, their contexts are relatively pure because only object or background pixels are shown in receptive fields except for few noise pixels.
Consequently, saliency networks can well classify these pixels without auxiliary techniques.
However, pixels located at the boundary between background and salient objects are so ambiguous that even experienced people are difficult to determine their labels.
From the perspective of features, these vectors extracted from motley image pixels fall near the hyperplanes, acting as hard examples.
As general saliency networks only apply pixel-wise binary classification, while neglect the boundary cues and train all pixels equally by cross entropy loss, they usually predict broad outline of target objects but are inferior in precise boundaries.

Base on the above observations, we argue that border pixels, as well as the hard examples in saliency maps, deserve much higher attention in the training phase.
Inspired by focal loss \cite{focalloss}, assigning higher weights to focus on these hard examples is theoretically and technologically convincing.
Towards this end, we apply spatial weight maps in cross entropy loss, which assigns relatively high value to emphasize pixels near the salient object borders.
The spatial weight map $M^{C}$ can be formulated as:
\begin{equation}
M^{C} = Guass(K\cdot((Y;S)^{+}-(Y;S)^{-}))+\mathds{1}
\end{equation}
where $(\star;S)^{+}$ and $(\star;S)^{-}$ represent dilation and erosion operations with the $5\times5$ mask $S$ respectively.
The object boundaries can be obtained by the difference between dilated and eroded images.
$K$ is a hyperparameter for assigning the high value which is set to 5 empirically.
To endow the pixels which are closed but not located at the boundaries with a moderate weight, we also adopt Guass function with a $5\times5$ range.
$\mathds{1}$ denotes the ones matrix with $224\times224$ resolution to set the pixels which is aloof from object boundaries to 1.
Compared with some boundary operators, such as Laplace operator, the above approach can generate thicker object contours for considerable error rates.

Generally, the proposed Contour Loss $L^{C}$ is implemented as the following formula:
\begin{equation}
   L^{C} = -\sum_{x, y}M^{C}_{x,y}\cdot(Y_{x,y}\cdot\log Y^{*}_{x,y}+(1-Y_{x,y})\cdot\log (1-Y^{*}_{x,y}))
\end{equation}
where $M^{C}_{x,y}$, $Y^{*}_{x,y}$ and $Y_{x,y}$ represent the spatial weight map, annotation map and predicted saliency map of the pixel $(x, y)$ respectively.
In implementation, since our network outputs multiple intermediate saliency maps, Contour Loss is applied to all intermediate maps to supervise network in the training process.
In other words, as we adopt Contour Loss, the $loss$ in Eq \ref{eq:loss} represents $L^{C}$.

\begin{figure}[!t]
\centering
\includegraphics[width=3.2in,height=2in]{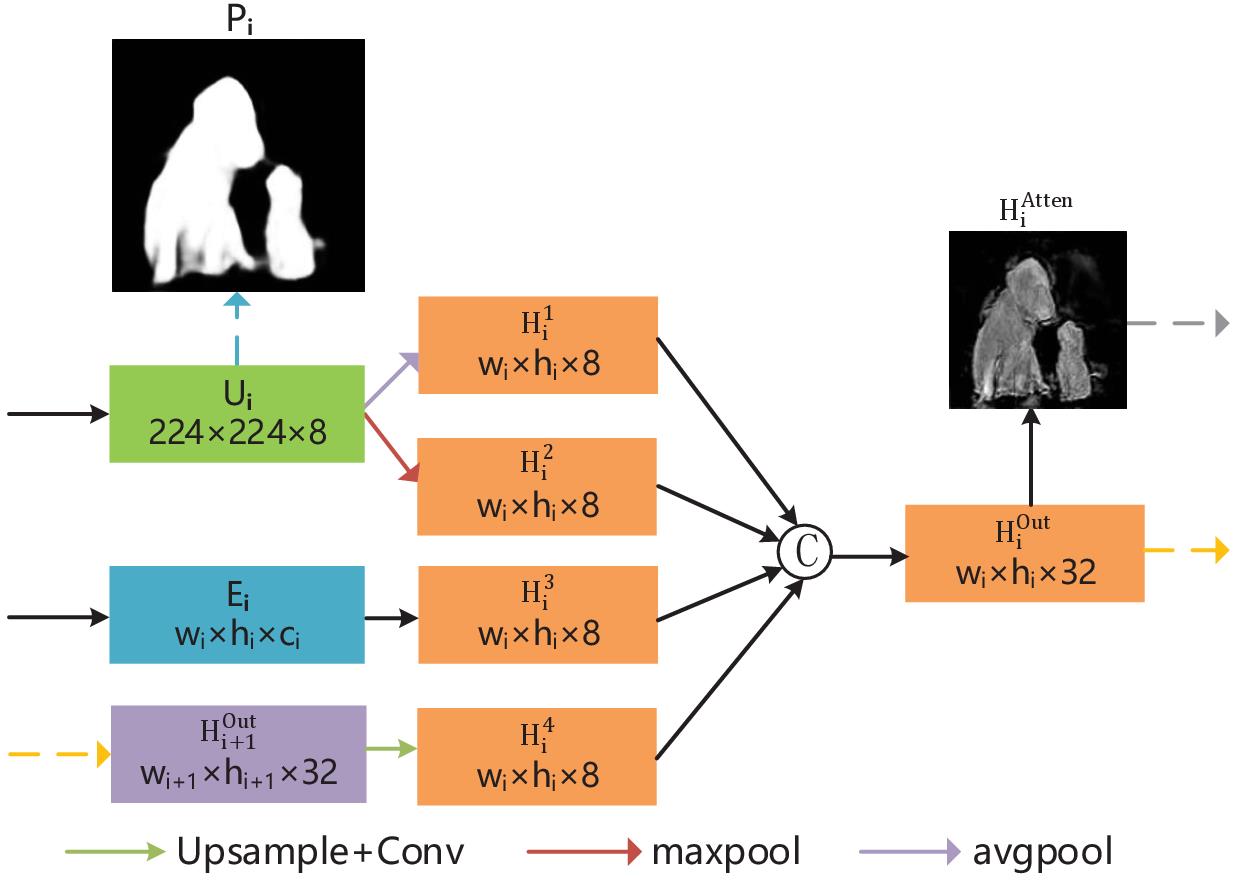}
\caption{Inner structure of the $i$th HGAM. 
Yellow, blue and gray dotted line represent attention stream, prediction stream and attention guidance respectively. 
$w_{i}$ and $h_{i}$ severally denote width and height of $E_{i}$. 
$H^{Out}_{i}$ and $H^{Atten}_{i}$ are the $i$th HGAM message and attention map respectively.
}
\label{fig:HGAM}
\end{figure}

\begin{table*}[!t]
\caption{
Quantitative comparisons of different saliency models on six benchmark datasets in terms of maximum $F_{\beta}$-measure and $MAE$ which are marked as $F^{*}_{\beta}$ and $mae$ in this table.
\textcolor{red}{Red}, \textcolor{blue}{blue} and \textcolor{green}{green} text indicate the best, second best and third best performance respectively. 
The computation speed (fps) are obtained on an NVIDIA TITAN X GPU.
}
\centering
\begin{tabular}{l|c|cc|cc|cc|cc|cc|cc}
\hline
 &\multirow{2}{*}{fps}&\multicolumn{2}{c|}{SOD} &\multicolumn{2}{c|}{PASCAL-S} &\multicolumn{2}{c|}{ECSSD} &\multicolumn{2}{c|}{HKU-IS} &\multicolumn{2}{c|}{DUTS-TE} &\multicolumn{2}{c}{DUT-O} \\\cline{3-14}
 & & $F^{*}_{\beta}$ &$mae$ &$F^{*}_{\beta}$ &$mae$ &$F^{*}_{\beta}$ &$mae$ &$F^{*}_{\beta}$ &$mae$ &$F^{*}_{\beta}$ &$mae$ &$F^{*}_{\beta}$ &$mae$ \\ \hline
\multicolumn{14}{c}{conventional methods} \\\hline
MR\cite{DUT-OMRON}   &1.1   &0.584 &0.237    &0.597 &0.209    &0.677 &0.173     &0.620 &0.180     &0.490 &0.220     &0.516 &0.210  \\
DRFI\cite{DRFI} &1.6    &0.697 &0.223    &0.698 &0.207    &0.786 &0.164     &0.777 &0.145     &0.647 &0.175     &0.690 &0.108  \\\hline
\multicolumn{14}{c}{ResNet-50 \cite{resnet} backbone} \\\hline
SRM\cite{SRM}  &14     &0.845 &0.132    &\textcolor{green}{0.862} &0.098    &0.917 &0.054     &0.906 &0.046     &0.827 &0.059     &0.769 &0.069 \\
BRN\cite{BRN}  &5.9     &\textcolor{blue}{0.858} &\textcolor{blue}{0.104}    &0.861 &\textcolor{blue}{0.071}       &0.921    &\textcolor{green}{0.045}     &0.916 &\textcolor{blue}{0.037}     &0.829 &\textcolor{green}{0.051}     &0.790 &\textcolor{green}{0.063}\\ \hline
\multicolumn{14}{c}{VGG-19 \cite{vgg} backbone} \\\hline
PAGRN\cite{PAGRN}  &--    &\centering-- &\centering--       &0.861 &0.092    &0.921  &0.064     &\textcolor{blue}{0.922} &0.048     &\textcolor{blue}{0.857} &0.055     &\textcolor{blue}{0.806} &0.072\\\hline 
\multicolumn{14}{c}{VGG-16 \cite{vgg} backbone} \\\hline
RFCN\cite{RFCN}    &9    &0.807 &0.166    &0.850 &0.132    &0.898 &0.095     &0.898 &0.080     &0.783 &0.090     &0.738 &0.095 \\
Amulet\cite{Amulet}  &16  &0.808 &0.145    &0.828 &0.103    &0.915 &0.059     &0.896 &0.052     &0.778 &0.085     &0.743 &0.098 \\
UCF\cite{UCF}     &23     &0.803 &0.169    &0.846 &0.128    &0.911 &0.078     &0.886 &0.074     &0.771 &0.117     &0.735 &0.132 \\
NLDF\cite{NLDF}   &12     &0.842 &0.130    &0.845 &0.112    &0.905 &0.063     &0.902 &0.048     &0.812 &0.066     &0.753 &0.080 \\
RA\cite{RA}       &35     &0.844 &0.124    &0.834 &0.104    &0.918 &0.059     &0.913 &0.045     &0.826 &0.055        &0.786 &\textcolor{blue}{0.062} \\
CKT\cite{CKT}     &23     &0.829 &0.119    &0.846 &0.081    &0.910 &0.054     &0.896 &0.048     &0.807 &0.062     &0.757 &0.071\\
BMP\cite{BMP}      &22 &0.851 &\textcolor{green}{0.106}    &\textcolor{green}{0.862} &\textcolor{green}{0.074}    &\textcolor{green}{0.928} &\textcolor{blue}{0.044}     &0.920 &\textcolor{green}{0.038}     &0.850 &\textcolor{blue}{0.049}     &\centering--    &{\centering--}    \\ 
PCA\cite{PiCANet}  &5.6 &\textcolor{green}{0.855} &0.108    &\textcolor{blue}{0.873} &0.088    &\textcolor{blue}{0.931} &0.047     &\textcolor{green}{0.921} &0.042     &\textcolor{green}{0.851} &0.054     &\textcolor{green}{0.794} &0.068 \\
Ours &26        &\textcolor{red}{0.873} &\textcolor{red}{0.095}    &\textcolor{red}{0.883} &\textcolor{red}{0.063}    &\textcolor{red}{0.933} &\textcolor{red}{0.037}     &\textcolor{red}{0.932} &\textcolor{red}{0.031}     &\textcolor{red}{0.872} &\textcolor{red}{0.042}     &\textcolor{red}{0.825} &\textcolor{red}{0.058} \\\hline
\end{tabular}
\label{metric comparison}
\end{table*}

\subsection{Hierarchical Global Attention Module}
The aim of salient object segmentation is to detect evident object regions, in other words, remove insignificant regions.
Although an original picture may contain multi-objects, not all the objects are conspicuous for saliency maps.
Thus, negligible information can lead to a sub-optimal result by distracting the models from salient regions.
To alleviate distractions of background, attention mechanism is a useful auxiliary module for salient object segmentation.
Since attention module can leverage the contextual information to generate a weight map, this map can guide the model to abate the insignificant features.
However, existing attention modules often adopt softmax function, which enormously emphasizes several important pixels and endows the others with a very small value.
Therefore these attention modules cannot attend to global contexts in high-resolution, which easily lead to overfitting in training.

Due to the above observations, instead of using softmax-wise attention modules, we utilize a novel function which is based on global contrast to attend to global contextual information.
Since a region is conspicuous in feature maps, each pixel in the region is also significant with a relatively large value, for example, over the mean.
In other words, the inconsequential features often have a relatively small value in feature maps, which are often smaller than the mean.
Thus, an intuitive method to abnegate the insignificant features is pixel-wisely subtracting the average value from feature maps.
After the subtraction, we can conduct a pixel-wise classification in feature maps: the positive pixels represent significant features while the negative ones denote inconsequential features.
Accordingly, the attention map $H^{Atten}$ can be generated as:
\begin{equation}
H^{Atten} = \delta(\frac{F^{In}-Aver(F^{In})}{\sqrt{Var(F^{In})+\epsilon}})+\lambda\mathds{1}
\label{eq:attention}
\end{equation}
where $F^{In}$ is the input feature map, $Aver$ and $Var$ represent the average and variance value of $F^{In}$ respectively.
$\lambda$ denotes a regularization term which is set to 0.1 empirically, while $\epsilon$ is a small value to avoid zero-division as the default setting.
Compared with softmax results, the pixel-wise disparity of our attention maps is more reasonable, in other words, our attention method can retain conspicuous regions from feature maps in high-resolution.

Since attention maps do not hold the labels, they are usually generated or supervised by predicted maps or ground truth masks which only retains the salient regions.
However, as models may also need information from background regions to determine salient objects, these attention maps may miss some crucial information.
In contrast to these attention modules like \cite{RA}, we exploit the feature maps which are near the predictions to generate unsupervised attention maps.
Therefore, our attention maps can not only leverage the strong feedbacks of supervised information to update themselves, but also contain the background information which may be crucial for saliency determination.

Towards this end, we propose our hierarchical global attention module (HGAM) to capture the multi-scale global contexts.
As shown in Fig \ref{fig:HGAM}, for a given HGAM $H_{i}$, it receives three inputs: the encoded feature $E_{i}$, the upsampled feature $U_{i}$ which is near the prediction $P_{i}$, and the previous HGAM message $H^{Out}_{i+1}$.
To extract the global contextual information from input features, we adopt maxpool and avgpool layers to deal with $U_{i}$ for obtaining contextual features $H^{1}_{i}$ and $H^{2}_{i}$ respectively, which is suggested by \cite{2pools}.
We also make channel-wise compression of $E_{i}$ to obtain $H^{3}_{i}$, while $H^{4}_{i}$ can be generated by the previous HGAM message $H^{Out}_{i+1}$ as:
\begin{equation}
H^{4}_{i}=
\begin{cases}
\delta(\{H^{Out}_{i+1}\}^{up\times2}; \theta^{H_{i}}_{4}) &2\le i<5 \\
\{\delta(maxpool(E_{5}; \theta^{H_{i}}_{4}))\}^{up\times2} &i=5
\end{cases}
\end{equation}
After obtaining $H^{1}_{i}$, ..., $H^{4}_{i}$, we make channel-wise concatenation of them to generate the $i$th HGAM message $H^{Out}_{i}$.
The $i$th attention map $H^{Atten}_{i}$ also can be generated by Eq \ref{eq:attention} with $H^{Out}_{i}$ as input feature.
Since $H_{i}$ obtain these two outputs, $H^{Out}_{i}$ is transferred to next HGAM $H_{i-1}$, while $H^{Atten}_{i}$ is utilized to guide the $Res_{i}$ as:
\begin{equation}
Res^{G}_{i}=H^{Atten}_{i}\odot Res_{i}
\end{equation}
where $Res^{G}_{i}$ is the guided feature map and $\odot$ represents the element-wise multiplication.

Different from the baseline network utilizes $P_{1}$ as final output, since $H^{Atten}_{1}$ guides $Res_{1}$ by recursively aggregating the multi-scale HGAM messages, we exploit the guided feature $Res^{G}_{1}$ to generate the final prediction $P$, which is also included in $F^{P}$.
Therefore, in training, $Loss$ in Eq \ref{eq:loss} can be rewrote as:
\begin{equation}
Loss(F^{P};Y)=loss(P;Y)+\sum_{i=1}^{5}{W_{i}^{L}\cdot loss(P_{i};Y)}
\label{eq:newloss}
\end{equation}
In testing, we adopt $P$ as our final prediction to evaluate our model.

\begin{figure*}[!t]
\centering
\begin{minipage}{0.24 \textwidth}
\includegraphics[width=1.8in]{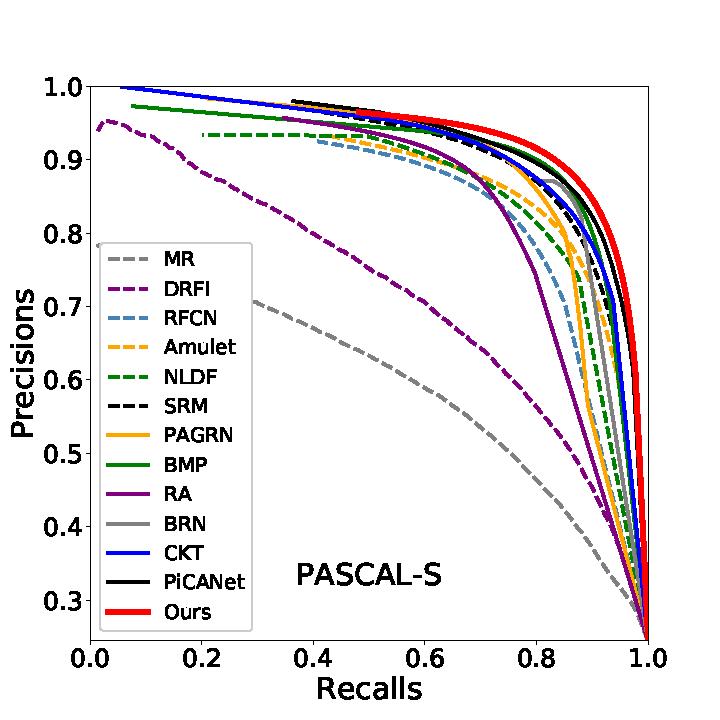}
\centering
\end{minipage}
\begin{minipage}{0.24 \textwidth}
\includegraphics[width=1.8in]{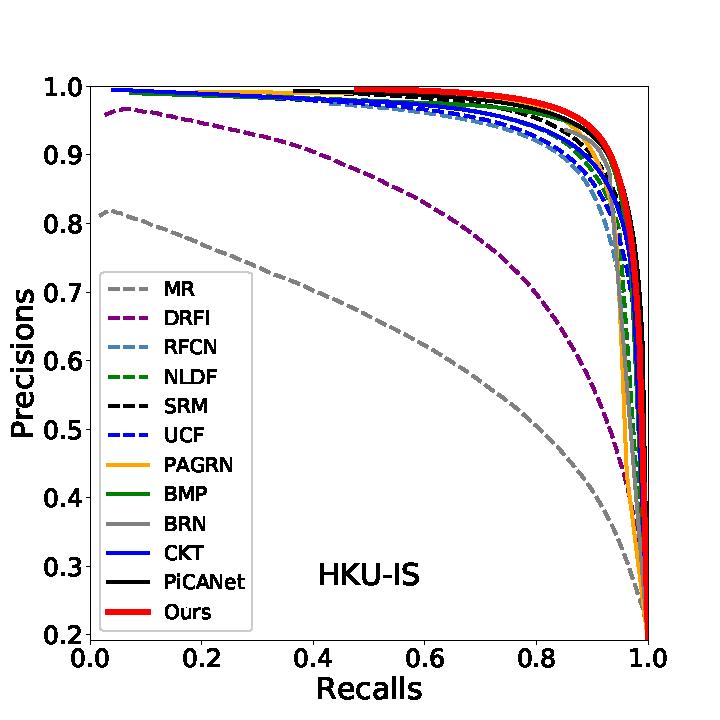}
\centering
\end{minipage}
\begin{minipage}{0.24 \textwidth}
\includegraphics[width=1.8in]{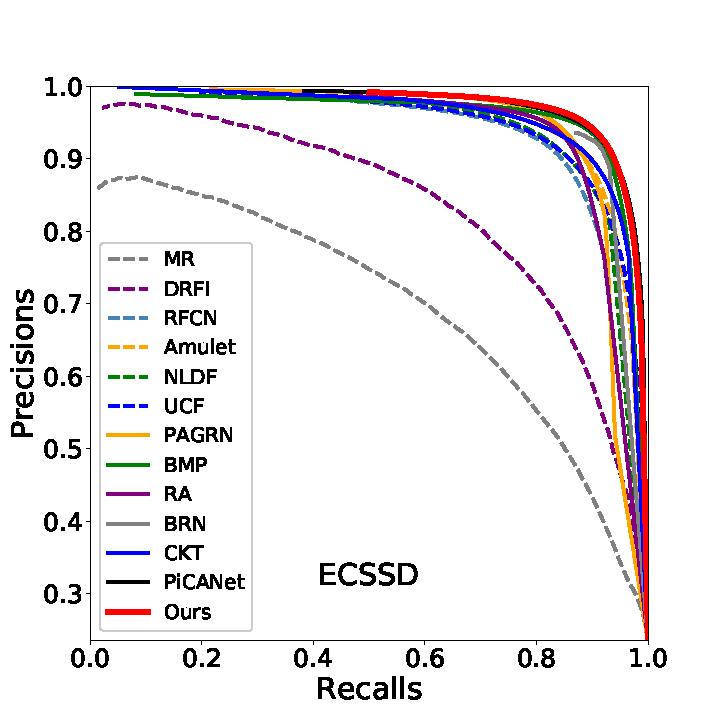}
\centering
\end{minipage}
\begin{minipage}{0.24 \textwidth}
\includegraphics[width=1.8in]{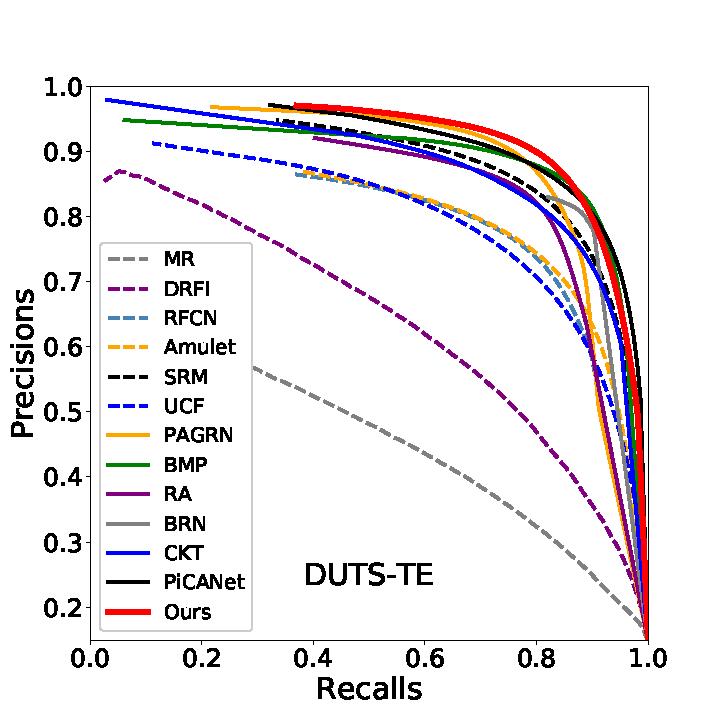}
\centering
\end{minipage}
\vspace{2pt}
\caption{PR curves of ours and other state-of-the-art methods.}
\label{fig:prcurve}
\end{figure*}

\section{Experiments}
\subsection{Experimental Settings}

\paragraph{Evaluation Datasets.}
To evaluate the performance of our model, six public saliency segmentation datasets are exploited.
\textbf{DUTS} \cite{DUTS} is a large scale saliency benchmark dataset which contains 10,553 images as trainging set (DUTS-TR) and 5,019 images as testing set (DUTS-TE).
In the experiments, we adopt DUTS-TR to train our model and DUTS-TE for evaluation.
For comprehensive evaluation, we also utilize \textbf{SOD} \cite{SOD}, \textbf{PASCAL-S} \cite{PASCAL-S}, \textbf{ECSSD} \cite{ECSSD}, \textbf{HKU-IS} \cite{HKU-IS} and \textbf{DUT-O} \cite{DUT-OMRON} for testing, which contain 300, 850, 1,000, 4,447 and 5,168 images respectively.
Note that for the testing on the abovementioned databases, no corresponding fine-tuning is carried.

\vspace{-0.4cm}
\paragraph{Implementation Details.} 
Our experiments are based on the Pytorch \cite{Pytorch} framework and run on a PC machine with a single NVIDIA TITAN X GPU (with 12G memory).

For training, we adopt DUTS-TR as training set and utilize data augmentation, which resamples each image to $256\times256$ before random flipping, and randomly crops the $224\times224$ region. 
We employ stochastic gradient descent (SGD) as the optimizer with a momentum (0.9) and a weight decay (1e-4). 
We also set basic learning rate to 1e-3 and finetune the VGG-16 \cite{vgg} backbone with a 0.05 times smaller learning rate. 
Since the saliency maps of hierarchical predictions are coarse to fine from $P_{5}$ to $P_{1}$, we set the incremental weights with these predictions.
Therefore $W_{5}^{L}$, ..., $W_{1}^{L}$ are set to 0.3, 0.4, 0.6, 0.8, 1 respectively in both Eq \ref{eq:loss} and \ref{eq:newloss}.
The minibatch size of our network is set to 10.
The maximum iteration is set to 150 epochs with the learning rate decay by a factor of 0.05 for each 10 epochs.
As it costs less than 500s for one epoch including training and evaluation, the total training time is below 21 hours.

For testing, follow the training settings, we also resize the feeding images to $224\times224$, and only utilize the final output $P$.
Since the testing time for each image is 0.038s, our model achieves 26 fps speed with $224\times224$ resolution.

\vspace{-0.4cm}
\paragraph{Evaluation Metrics.}
To evaluate different algorithms, we adopt three metrics for the quality of saliency maps, including the precision-recall (PR) curves, $F_{\beta}$-measure \cite{F-score} and mean absolute error ($MAE$).

To evaluate the robustness of saliency results in different thresholds, we utilize the PR curve to demonstrate the relation of precision and recall by thresholding saliency maps from 0 to 255.

The $F_{\beta}$-measure is a weighted combination of precision and recall value for saliency maps, which can be calculated by
\begin{equation}
F_{\beta}=\frac{(1+\beta^{2})\times Precision\times Recall}{\beta^{2}\times Precision+Recall}
\end{equation}
where $\beta^{2}$ is set to 0.3 as suggested in \cite{F-score}.
To alleviate the unfairness caused by different thresholds in papers, we report the maximum $F_{\beta}$-measure as suggested by \cite{NLDF,BMP}, which selects the best score over all thresholds from 0 to 255.

For comprehensive comparisons, we also adopt the $MAE$ metric to evaluate the pixel-wise average absolute difference between the saliency map $S$ and its corresponding ground truth mask $G$,
\begin{equation}
MAE = \frac{1}{w\times h}\sum^{w}_{i=1}\sum^{h}_{j=1}{|S_{ij}-G_{ij}|}
\end{equation}
where $w$ and $h$ represent the width and height of a given picture respectively.

\subsection{Comparison with State-of-the-arts}
To evaluate the performance, we compare our method with 13 state-of-the-art algorithms on aforementioned six public benchmarks in terms of visual evaluation, PR curve, maximum $F_{\beta}$-measure and $MAE$ metrics.
These methods include 2 conventional algorithms: DRFI \cite{DRFI}, MR \cite{DUT-OMRON}, as well as 11 deep learning models: RFCN \cite{RFCN}, Amulet \cite{Amulet}, UCF \cite{UCF}, NLDF \cite{NLDF}, SRM \cite{SRM}, PAGRN \cite{PAGRN}, BRN \cite{BRN}, CKT \cite{CKT}, BMP \cite{BMP}, PCA \cite{PiCANet} and RA \cite{RA}.

\begin{figure}[!t]
\centering
\begin{minipage}{0.155 \textwidth}
\includegraphics[width=1.08in, height=0.8in]{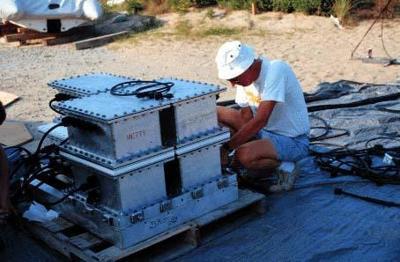}
\end{minipage}
\begin{minipage}{0.155 \textwidth}
\includegraphics[width=1.08in, height=0.8in]{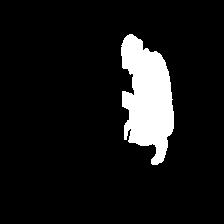}
\end{minipage}
\begin{minipage}{0.155 \textwidth}
\includegraphics[width=1.08in, height=0.8in]{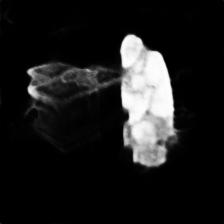}
\end{minipage}
\begin{minipage}{0.155 \textwidth}
\centering (a) image
\end{minipage}
\begin{minipage}{0.155 \textwidth}
\centering (b) gt
\end{minipage}
\begin{minipage}{0.155 \textwidth}
\centering (c) ours($\mathcal{B}$+$\mathcal{C}$+$\mathcal{H}$) 
\end{minipage}

\begin{minipage}{0.155 \textwidth}
\includegraphics[width=1.08in, height=0.8in]{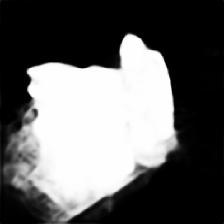}
\end{minipage}
\begin{minipage}{0.155 \textwidth}
\includegraphics[width=1.08in, height=0.8in]{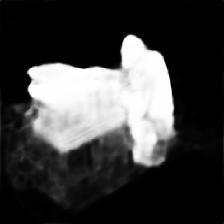}
\end{minipage}
\begin{minipage}{0.155 \textwidth}
\includegraphics[width=1.08in, height=0.8in]{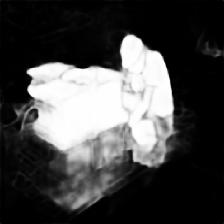}
\end{minipage}
\begin{minipage}{0.155 \textwidth}
\centering (d) $\mathcal{B}$
\end{minipage}
\begin{minipage}{0.155 \textwidth}
\centering (e) $\mathcal{B}$ + $\mathcal{H}$
\end{minipage}
\begin{minipage}{0.155 \textwidth}
\centering (f) $\mathcal{B}$ + $\mathcal{C}$
\end{minipage}

\caption{Visual examples under different settings.
$\mathcal{B}$, $\mathcal{C}$ and $\mathcal{H}$ denote baseline network, Contour Loss and HGAM respectively.
}
\label{fig:ablation}
\end{figure}

\begin{figure*}[!t]
\centering
\begin{minipage}{0.095 \textwidth}
\includegraphics[width=0.67in]{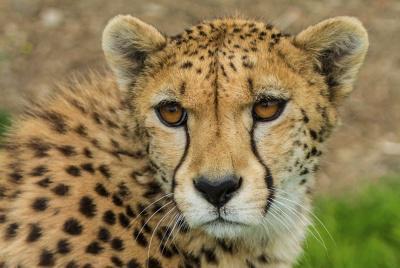}
\end{minipage}
\begin{minipage}{0.095 \textwidth}
\includegraphics[width=0.67in]{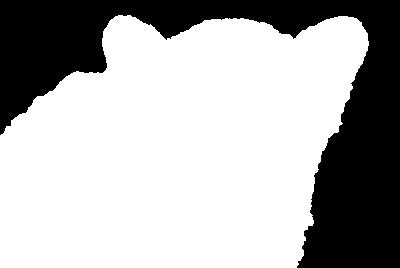}
\end{minipage}
\begin{minipage}{0.095 \textwidth}
\includegraphics[width=0.67in]{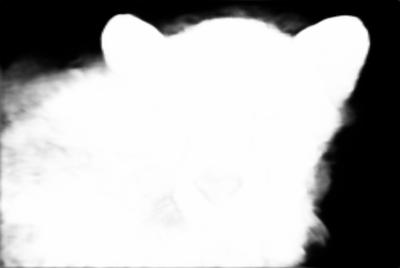}
\end{minipage}
\begin{minipage}{0.095 \textwidth}
\includegraphics[width=0.67in]{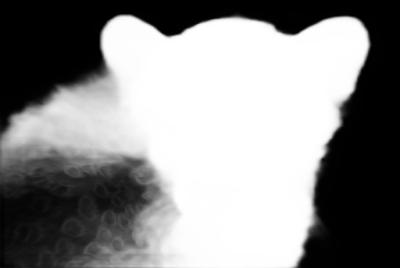}
\end{minipage}
\begin{minipage}{0.095 \textwidth}
\includegraphics[width=0.67in]{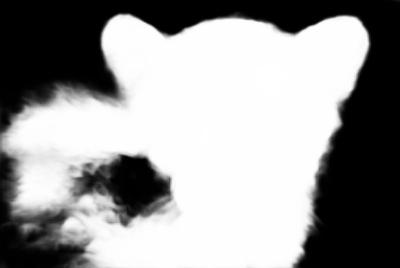}
\end{minipage}
\begin{minipage}{0.095 \textwidth}
\includegraphics[width=0.67in]{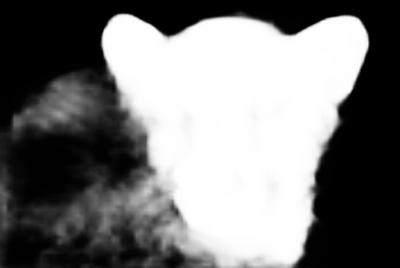}
\end{minipage}
\begin{minipage}{0.095 \textwidth}
\includegraphics[width=0.67in]{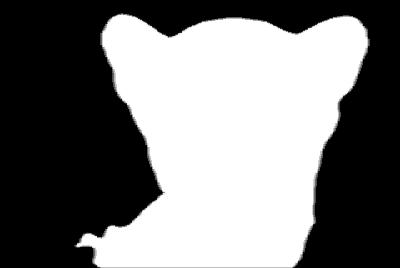}
\end{minipage}
\begin{minipage}{0.095 \textwidth}
\includegraphics[width=0.67in]{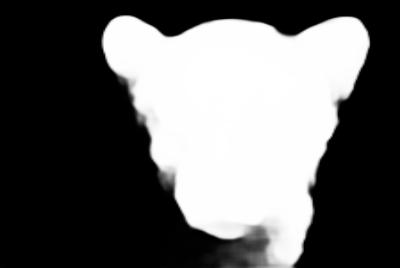}
\end{minipage}
\begin{minipage}{0.095 \textwidth}
\includegraphics[width=0.67in]{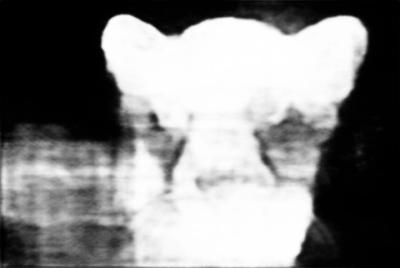}
\end{minipage}
\begin{minipage}{0.095 \textwidth}
\includegraphics[width=0.67in]{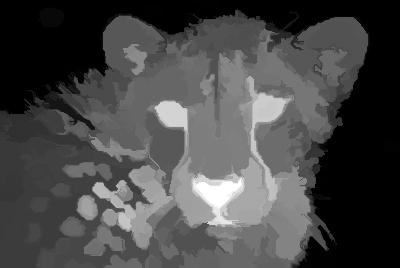}
\end{minipage}

\begin{minipage}{0.095 \textwidth}
\includegraphics[width=0.67in]{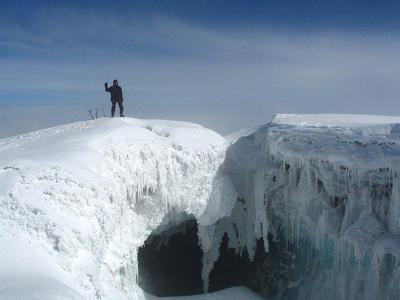}
\end{minipage}
\begin{minipage}{0.095 \textwidth}
\includegraphics[width=0.67in]{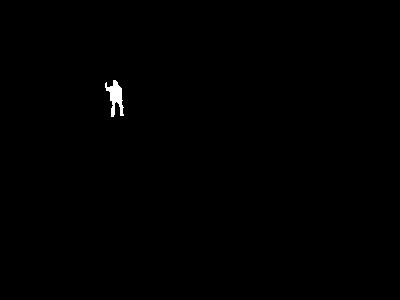}
\end{minipage}
\begin{minipage}{0.095 \textwidth}
\includegraphics[width=0.67in]{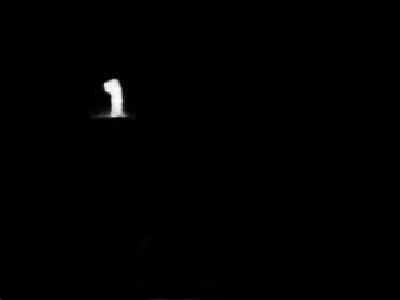}
\end{minipage}
\begin{minipage}{0.095 \textwidth}
\includegraphics[width=0.67in]{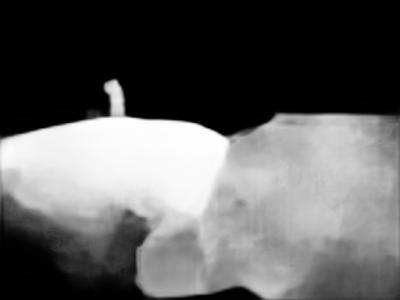}
\end{minipage}
\begin{minipage}{0.095 \textwidth}
\includegraphics[width=0.67in]{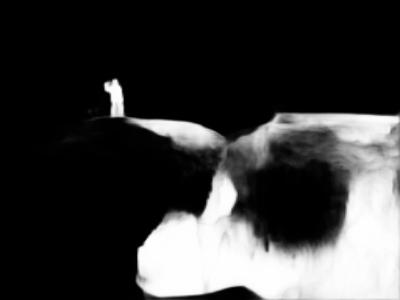}
\end{minipage}
\begin{minipage}{0.095 \textwidth}
\includegraphics[width=0.67in]{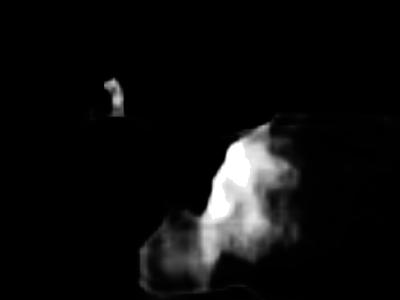}
\end{minipage}
\begin{minipage}{0.095 \textwidth}
\includegraphics[width=0.67in]{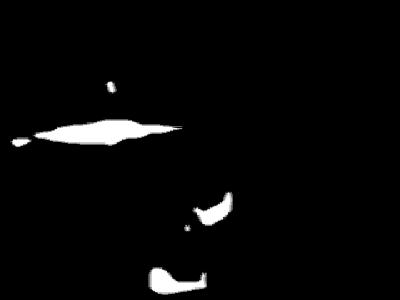}
\end{minipage}
\begin{minipage}{0.095 \textwidth}
\includegraphics[width=0.67in]{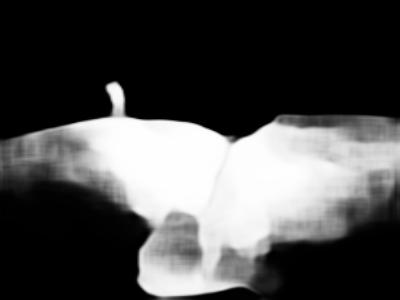}
\end{minipage}
\begin{minipage}{0.095 \textwidth}
\includegraphics[width=0.67in]{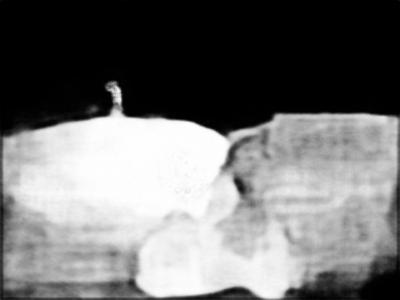}
\end{minipage}
\begin{minipage}{0.095 \textwidth}
\includegraphics[width=0.67in]{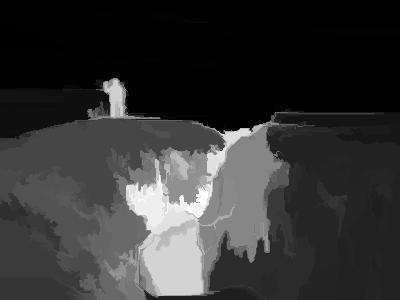}
\end{minipage}

\begin{minipage}{0.095 \textwidth}
\includegraphics[width=0.67in]{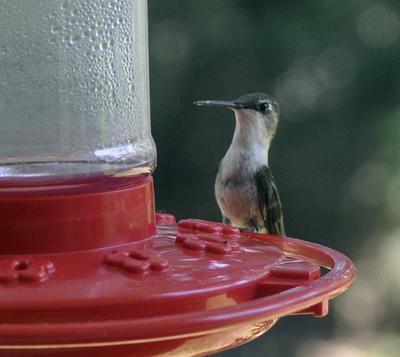}
\end{minipage}
\begin{minipage}{0.095 \textwidth}
\includegraphics[width=0.67in]{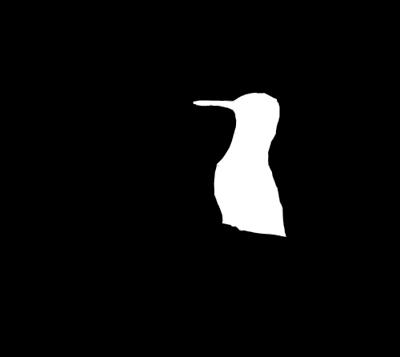}
\end{minipage}
\begin{minipage}{0.095 \textwidth}
\includegraphics[width=0.67in]{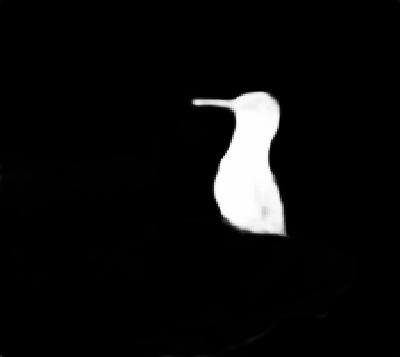}
\end{minipage}
\begin{minipage}{0.095 \textwidth}
\includegraphics[width=0.67in]{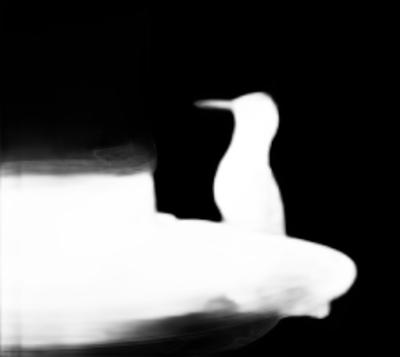}
\end{minipage}
\begin{minipage}{0.095 \textwidth}
\includegraphics[width=0.67in]{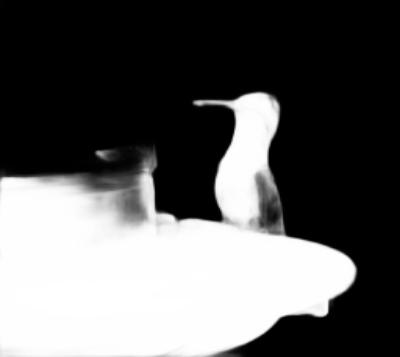}
\end{minipage}
\begin{minipage}{0.095 \textwidth}
\includegraphics[width=0.67in]{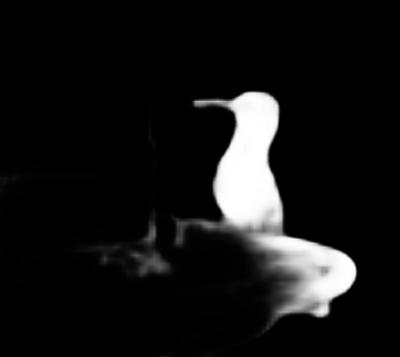}
\end{minipage}
\begin{minipage}{0.095 \textwidth}
\includegraphics[width=0.67in]{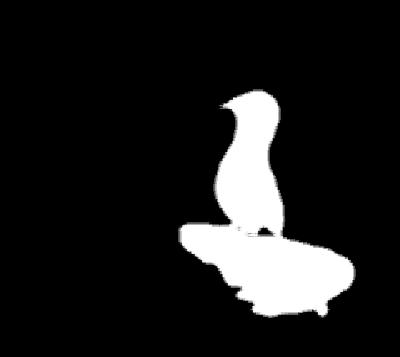}
\end{minipage}
\begin{minipage}{0.095 \textwidth}
\includegraphics[width=0.67in]{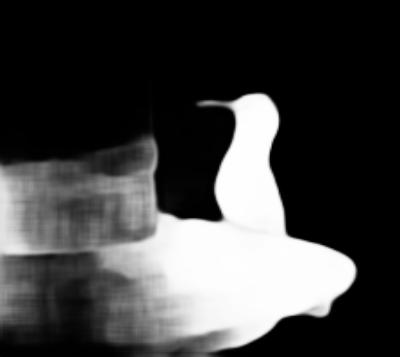}
\end{minipage}
\begin{minipage}{0.095 \textwidth}
\includegraphics[width=0.67in]{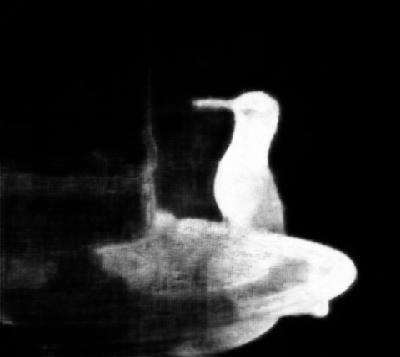}
\end{minipage}
\begin{minipage}{0.095 \textwidth}
\includegraphics[width=0.67in]{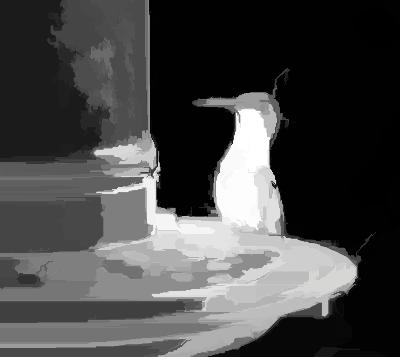}
\end{minipage}

\begin{minipage}{0.095 \textwidth}
\includegraphics[width=0.67in]{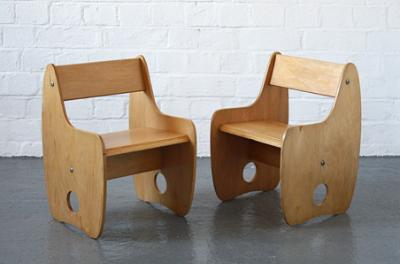}
\end{minipage}
\begin{minipage}{0.095 \textwidth}
\includegraphics[width=0.67in]{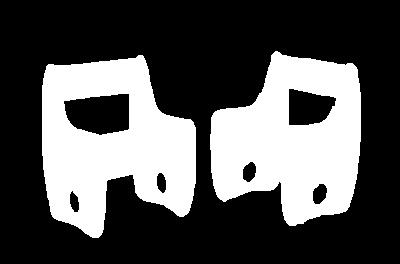}
\end{minipage}
\begin{minipage}{0.095 \textwidth}
\includegraphics[width=0.67in]{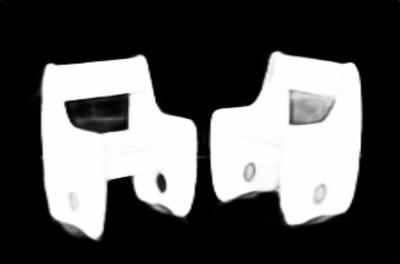}
\end{minipage}
\begin{minipage}{0.095 \textwidth}
\includegraphics[width=0.67in]{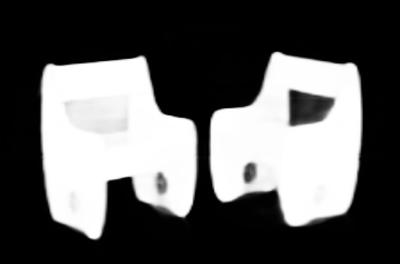}
\end{minipage}
\begin{minipage}{0.095 \textwidth}
\includegraphics[width=0.67in]{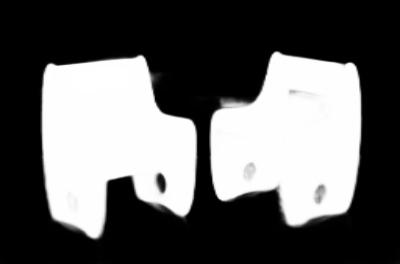}
\end{minipage}
\begin{minipage}{0.095 \textwidth}
\includegraphics[width=0.67in]{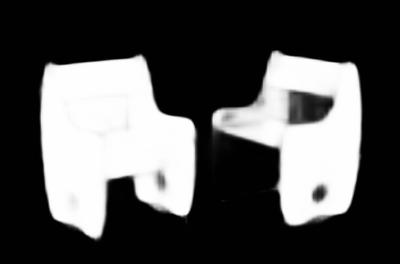}
\end{minipage}
\begin{minipage}{0.095 \textwidth}
\includegraphics[width=0.67in]{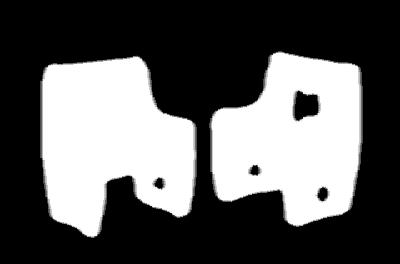}
\end{minipage}
\begin{minipage}{0.095 \textwidth}
\includegraphics[width=0.67in]{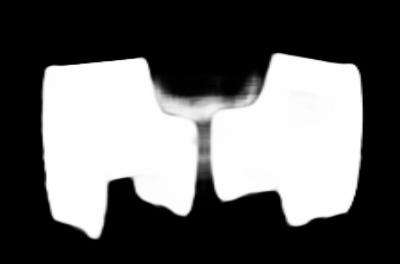}
\end{minipage}
\begin{minipage}{0.095 \textwidth}
\includegraphics[width=0.67in]{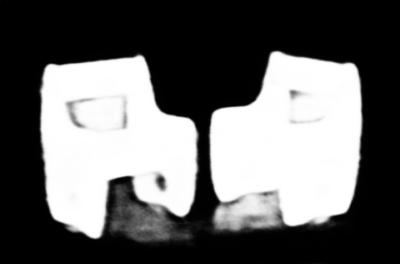}
\end{minipage}
\begin{minipage}{0.095 \textwidth}
\includegraphics[width=0.67in]{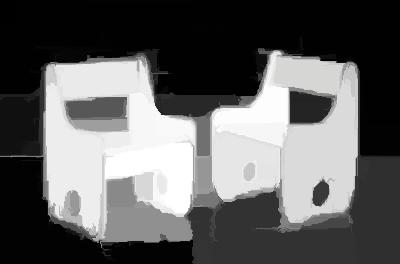}
\end{minipage}

\begin{minipage}{0.095 \textwidth}
\includegraphics[width=0.67in]{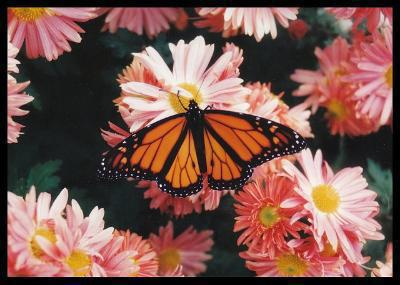}
\centering
\end{minipage}
\begin{minipage}{0.095 \textwidth}
\includegraphics[width=0.67in]{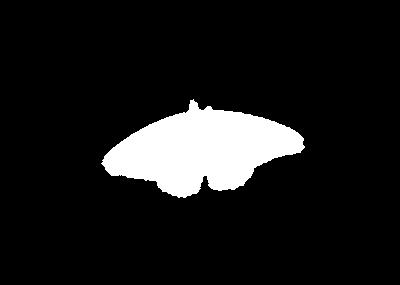}
\centering
\end{minipage}
\begin{minipage}{0.095 \textwidth}
\includegraphics[width=0.67in]{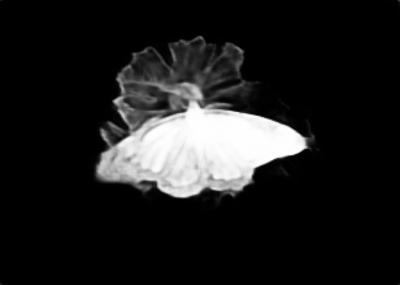}
\centering
\end{minipage}
\begin{minipage}{0.095 \textwidth}
\includegraphics[width=0.67in]{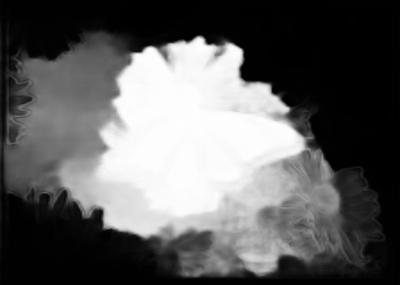}
\centering
\end{minipage}
\begin{minipage}{0.095 \textwidth}
\includegraphics[width=0.67in]{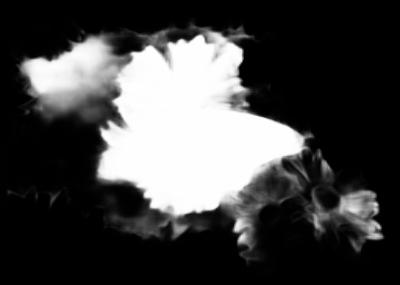}
\centering
\end{minipage}
\begin{minipage}{0.095 \textwidth}
\includegraphics[width=0.67in]{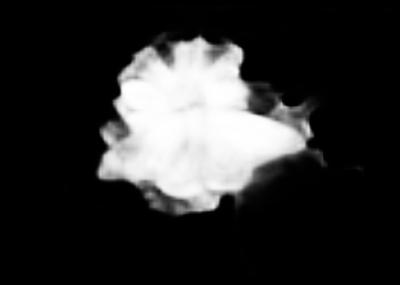}
\centering
\end{minipage}
\begin{minipage}{0.095 \textwidth}
\includegraphics[width=0.67in]{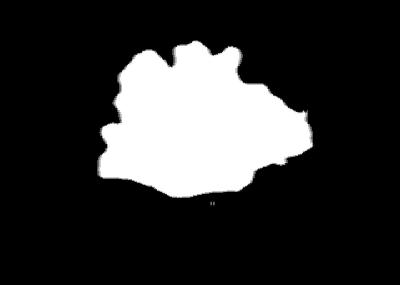}
\centering
\end{minipage}
\begin{minipage}{0.095 \textwidth}
\includegraphics[width=0.67in]{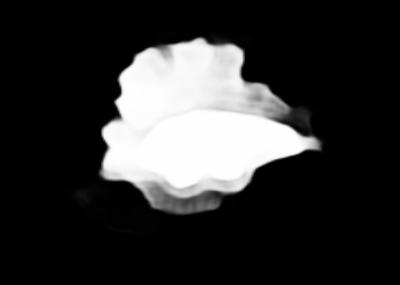}
\centering
\end{minipage}
\begin{minipage}{0.095 \textwidth}
\includegraphics[width=0.67in]{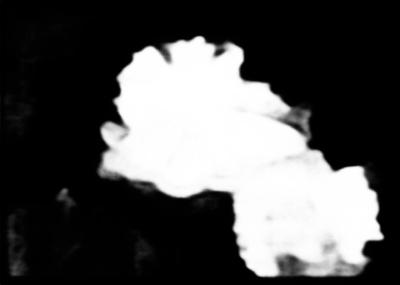}
\centering
\end{minipage}
\begin{minipage}{0.095 \textwidth}
\includegraphics[width=0.67in]{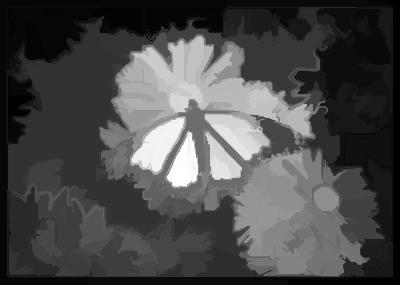}
\centering
\end{minipage}

\begin{minipage}{0.095 \textwidth}
\vspace{2pt}
\centering
image
\end{minipage}
\begin{minipage}{0.095 \textwidth}
\vspace{2pt}
\centering
gt
\end{minipage}
\begin{minipage}{0.095 \textwidth}
\vspace{2pt}
\centering
ours
\end{minipage}
\begin{minipage}{0.095 \textwidth}
\vspace{2pt}
\centering
PCA\cite{PiCANet}
\end{minipage}
\begin{minipage}{0.095 \textwidth}
\vspace{2pt}
\centering
BMP\cite{BMP}
\end{minipage}
\begin{minipage}{0.095 \textwidth}
\vspace{2pt}
\centering
PAGRN\cite{PAGRN}
\end{minipage}
\begin{minipage}{0.095 \textwidth}
\vspace{2pt}
\centering
BRN\cite{BRN}
\end{minipage}
\begin{minipage}{0.095 \textwidth}
\vspace{2pt}
\centering
CKT\cite{CKT}
\end{minipage}
\begin{minipage}{0.095 \textwidth}
\vspace{2pt}
\centering
UCF\cite{UCF}
\end{minipage}
\begin{minipage}{0.095 \textwidth}
\vspace{2pt}
\centering
DRFI\cite{resnet}
\end{minipage}
\vspace{2pt}
\caption{Visual comparison between ours and other state-of-the-art methods.}
\label{fig:results}
\end{figure*}

\vspace{-0.4cm}
\paragraph{Visual Comparison.}
The visual comparison between ours and other state-of-the-arts is shown in Fig \ref{fig:results}.
It can be observed that our method well detect the target objects in various situations, i.e., containing the object too huge or too small (rows 1 and 2), object touching image edges (row 1), object touching other inconsequential items (row 3), multi-objects (row 4) and object appearance similar with background (row 5).
It is also worth noting that our results have finer boundaries and more precise localization of salient regions, which thanks to the effect of Contour Loss and HGAM respectively.

\vspace{-0.4cm}
\paragraph{F-measure and MAE.}
In Table \ref{metric comparison}, we show quantitative evaluation results between ours and other superior methods under maximum $F_{\beta}$-measure and $MAE$ metrics.
To the best of our knowledge, as only utilize the VGG-16 backbone without any post-processing methods like CRF \cite{CRF}, our model surpasses all existing networks and significantly refresh state-of-the-art performance on benchmarks by 1 to 2 percent.


\vspace{-0.4cm}
\paragraph{PR curve.}
In Fig \ref{fig:prcurve}, we compare our approach with other state-of-the-art methods in terms of PR curve on 4 benchmarks.
It can be observed that our model consistently outperforms all the other methods.


\vspace{-0.4cm}
\paragraph{Computational Cost.}
Besides, Table \ref{metric comparison} also provides the average testing time for each image among the state-of-the-arts on an NVIDIA TITAN X GPU.
We can see that our approach only takes 0.038s (corresponding to 26fps) to generate a saliency map, which is faster than other mainstream methods.

\begin{figure}[!t]
\centering
\begin{minipage}{0.115 \textwidth}
\includegraphics[width=0.81in,height=0.65in]{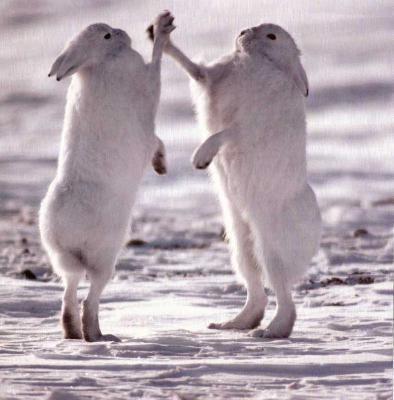}
\end{minipage}
\begin{minipage}{0.115 \textwidth}
\includegraphics[width=0.81in,height=0.65in]{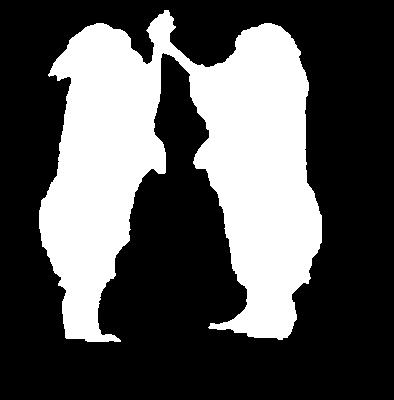}
\end{minipage}
\begin{minipage}{0.115 \textwidth}
\includegraphics[width=0.81in,height=0.65in]{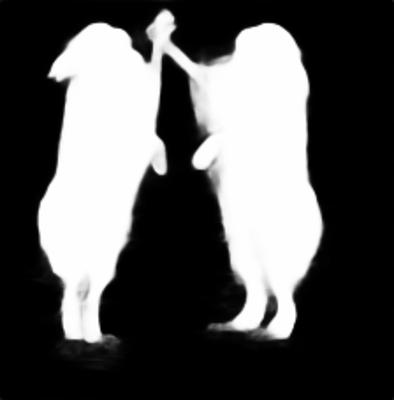}
\end{minipage}
\begin{minipage}{0.115 \textwidth}
\includegraphics[width=0.81in,height=0.65in]{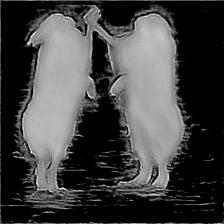}
\end{minipage}

\begin{minipage}{0.115 \textwidth}
\centering (a) image
\end{minipage}
\begin{minipage}{0.115 \textwidth}
\centering (b) gt
\end{minipage}
\begin{minipage}{0.115 \textwidth}
\centering (c) $P$
\end{minipage}
\begin{minipage}{0.115 \textwidth}
\centering (d) $H^{Atten}_{1}$
\end{minipage}

\begin{minipage}{0.115 \textwidth}
\includegraphics[width=0.81in,height=0.65in]{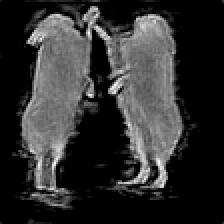}
\end{minipage}
\begin{minipage}{0.115 \textwidth}
\includegraphics[width=0.81in,height=0.65in]{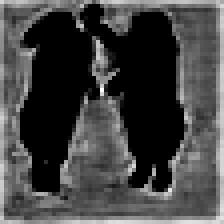}
\end{minipage}
\begin{minipage}{0.115 \textwidth}
\includegraphics[width=0.81in,height=0.65in]{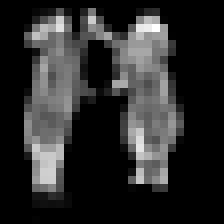}
\end{minipage}
\begin{minipage}{0.115 \textwidth}
\includegraphics[width=0.81in,height=0.65in]{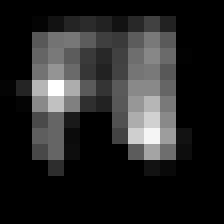}
\end{minipage}

\begin{minipage}{0.115 \textwidth}
\centering (e) $H^{Atten}_{2}$
\end{minipage}
\begin{minipage}{0.115 \textwidth}
\centering (f) $H^{Atten}_{3}$
\end{minipage}
\begin{minipage}{0.115 \textwidth}
\centering (g) $H^{Atten}_{4}$
\end{minipage}
\begin{minipage}{0.115 \textwidth}
\centering (h) $H^{Atten}_{5}$
\end{minipage}

\caption{Visualization of attention maps.}
\label{fig:visualization}
\end{figure}

\begin{table}[!t]
\caption{Comparison of different settings in terms of maximum $F_{\beta}$-measure and $MAE$ metric, which are marked as $F^{*}_{\beta}$ and $mae$ in this table.
$\mathcal{B}$, $\mathcal{C}$ and $\mathcal{H}$ represent the baseline network, Contour Loss and HGAM respectively.
\textbf{Bold} text indicates the best performance in table.}

\centering
\begin{tabular}{l|cc|cc}
\hline
                        &\multicolumn{2}{c|}{DUTS-TE\cite{DUTS}}  &\multicolumn{2}{c}{DUT-O\cite{DUT-OMRON}}   \\\cline{2-5}
                  &$F^{*}_{\beta}$         &$mae$                   &$F^{*}_{\beta}$         &$mae$    \\\hline
$\mathcal{B}$    &0.848                   &0.050                   &0.787                   &0.070    \\\hline 
$\mathcal{B}$+$\mathcal{C}$                  &0.861                   &0.044                   &0.806                   &0.063    \\\hline
$\mathcal{B}$+$\mathcal{H}$                  &0.860                   &0.048                   &0.801                   &0.069    \\\hline
ours($\mathcal{B}$+$\mathcal{C}$+$\mathcal{H}$)              &\textbf{0.872}          &\textbf{0.042}          &\textbf{0.825}          &\textbf{0.058}    \\\hline
\end{tabular}
\label{ablation}
\end{table}

\subsection{Ablation Study}
To evaluate the effectiveness of the proposed Contour Loss and HGAM, we show the results of quantitative and visual comparison under different settings.
Table \ref{ablation} shows the quantitative comparison which demonstrates that only utilizing Contour Loss or HGAM  can enhance the baseline performance by nearly 2 percent.
As incorporating Contour Loss and HGAM can make a further improvement on two massive datasets by 1 to 2 precent, which proves that Contour Loss and HGAM refine the saliency results from different aspects.

In Fig \ref{fig:ablation}, compared with the baseline result, Contour Loss can obtain a finer boundaries while HGAM is better in eliminating background distractions.
Since our result outperforms the other results both in boundaries and background elimination, it proves that incorporating Contour Loss and HGAM can lead to mutual promotion in training.

\subsection{HGAM Visualizaiton}
As shown in Fig \ref{fig:visualization}, we visualize the attention maps generated by HGAM to further understand how it works.
We can observe that these attention maps show fine-to-coarse locations of salient objects from $H^{Atten}_{1}$ to $H^{Atten}_{5}$, which greatly matches the global attention mechanism in different resolutions.
It is worth pondering that, different from other attention maps which only focus on salient regions, $H^{Atten}_{3}$ assigns a higher value to the regions which are corresponding to background.
As $P$ well abnegates these background regions, we reckon that the model also needs to perceive background regions for eliminating the insignificant features.
Moreover, as $H^{Atten}_{1}$ shows the clear boundaries of salient objects, it is convincingly proved that the mutual promotion of Contour Loss and HGAM in boundary-aware learning.

\section{Conclusion}
We propose the Contour Loss and the HGAM to help networks learn to better detect saliency objects in visual range.
The Contour Loss forces to learn boundary-wise distinctions between salient objects and background, while HGAM enables the models to capture global contextual information in all resolutions.
Experimental results on six datasets demonstrate that our proposed approach outperforms 13 state-of-the-art methods under different evaluation metrics.

{\small
\bibliographystyle{ieee}
\bibliography{egbib}
}

\end{document}